\documentclass[11pt,a4paper]{article}

\usepackage{multirow}
\usepackage{amsmath}
\usepackage[dvipsnames]{xcolor}
\usepackage{url}
\usepackage{array}
\usepackage{soul}
\usepackage{graphicx}
\usepackage{enumitem}
\usepackage{makecell}

\setlist{nolistsep}

\usepackage[hyperref]{emnlp-ijcnlp-2019}
\usepackage{times}
\usepackage{latexsym}

\usepackage{cleveref}

\aclfinalcopy 

\title{Dreaddit: A Reddit Dataset for Stress Analysis in Social Media}

\author{Elsbeth Turcan, Kathleen McKeown \\
  Columbia University \\
  Department of Computer Science \\
  {\tt \{eturcan, kathy\}@cs.columbia.edu}}

\date{}

\begin{document}
\maketitle
\begin{abstract}
  Stress is a nigh-universal human experience, particularly in the online world. While stress can be a motivator, too much stress is associated with many negative health outcomes, making its identification useful across a range of domains. However, existing computational research typically only studies stress in domains such as speech, or in short genres such as Twitter. We present Dreaddit, a new text corpus of lengthy multi-domain social media data for the identification of stress. Our dataset consists of 190K posts from five different categories of Reddit communities; we additionally label 3.5K total segments taken from 3K posts using Amazon Mechanical Turk. We present preliminary supervised learning methods for identifying stress, both neural and traditional, and analyze the complexity and diversity of the data and characteristics of each category.
\end{abstract}

\section{Introduction}

In our online world, social media users tweet, post, and message an incredible number of times each day, and the interconnected, information-heavy nature of our lives makes stress more prominent and easily observable than ever before. With many platforms such as Twitter, Reddit, and Facebook, the scientific community has access to a massive amount of data to study the daily worries and stresses of people across the world.\footnote{\url{https://www.gse.harvard.edu/news/uk/17/12/social-media-and-teen-anxiety}}

Stress is a nearly universal phenomenon, and we have some evidence of its prevalence and recent increase. For example, the American Psychological Association (APA) has performed annual studies assessing stress in the United States since 2007\footnote{\url{https://www.apa.org/news/press/releases/stress/index?tab=2}} which demonstrate widespread experiences of chronic stress. Stress is a subjective experience whose effects and even definition can vary from person to person; as a baseline, the APA defines stress as a reaction to extant and future demands and pressures,\footnote{\url{https://www.apa.org/helpcenter/stress-kinds}} which can be positive in moderation. Health and psychology researchers have extensively studied the connection between too much stress and physical and mental health~\citep{lupien_mcewen_gunnar_heim_2009,calcia_bonsall_bloomfield_selvaraj_barichello_howes_2016}. 

In this work, we present a corpus of social media text for detecting the presence of stress. We hope this corpus will facilitate the development of models for this problem, which has diverse applications in areas such as diagnosing physical and mental illness, gauging public mood and worries in politics and economics, and tracking the effects of disasters. Our contributions are as follows:

\begin{itemize}
    \item Dreaddit, a dataset of lengthy social media posts in five categories, each including stressful and non-stressful text and different ways of expressing stress, with a subset of the data annotated by human annotators;\footnote{Our dataset will be made available at \url{http://www.cs.columbia.edu/~eturcan/data/dreaddit.zip}.}
    \item Supervised models, both discrete and neural, for predicting stress, providing benchmarks to stimulate further work in the area; and
    \item Analysis of the content of our dataset and the performance of our models, which provides insight into the problem of stress detection.
\end{itemize}

In the remainder of this paper, we will review relevant work, describe our dataset and its annotation, provide some analysis of the data and stress detection problem, present and discuss results of some supervised models on our dataset, and finally conclude with our summary and future work.

\section{Related Work}

Because of the subjective nature of stress, relevant research tends to focus on physical signals, such as cortisol levels in saliva \citep{allen_kennedy_cryan_dinan_clarke_2014}, electroencephalogram (EEG) readings \citep{alshargie_kiguchi_badruddin_dass_hani_2016}, or speech data \citep{zuo-etal-2012-multilingual}. This work captures important aspects of the human reaction to stress, but has the disadvantage that hardware or physical presence is required. However, because of the aforementioned proliferation of stress on social media, we believe that stress can be observed and studied purely from text.

Other threads of research have also made this observation and generally use microblog data (e.g., Twitter). The most similar work to ours includes \citet{DBLP:journals/corr/abs-1805-12307}, who use Long Short-Term Memory Networks (LSTMs) to detect stress in speech and Twitter data; \citet{DBLP:journals/corr/abs-1811-07430}, who examine the Facebook and Twitter posts of users who score highly on a diagnostic stress questionnaire; and \citet{7885098}, who detect stress on microblogging websites using a Convolutional Neural Network (CNN) and factor graph model with a suite of discrete features. Our work is unique in that it uses data from Reddit, which is both typically longer and not typically as conducive to distant labeling as microblogs (which are labeled in the above work with hashtags or pattern matching, such as ``I feel stressed''). The length of our posts will ultimately enable research into the causes of stress and will allow us to identify more implicit indicators. We also limit ourselves to text data and metadata (e.g., posting time, number of replies), whereas \citet{DBLP:journals/corr/abs-1805-12307} also train on speech data and \citet{7885098} include information from photos, neither of which is always available. Finally, we label individual parts of longer posts for acute stress using human annotators, while \citet{DBLP:journals/corr/abs-1811-07430} label users themselves for chronic stress with the users' voluntary answers to a psychological questionnaire.

Researchers have used Reddit data to examine a variety of mental health conditions such as depression \citep{ICWSM136124} and other clinical diagnoses such as general anxiety \citep{cohan-etal-2018-smhd}, but to our knowledge, our corpus is the first to focus on stress as a general experience, not only a clinical concept.

\section{Dataset}

\subsection{Reddit Data}

\begin{figure}
    \centering
    \fbox{\parbox{7cm}{I have this \hl{feeling of dread} about school right before I go to bed and I wake up with an \hl{upset stomach} which lasts all day and nakes me \hl{feel like I'll throw up}. This causes me to \hl{lose appetite} and \hl{not wanting to drink water for fear of throwing up}. \hl{I'm not sure where else to go} with this, but \hl{I need help}. If any of you have this, can you tell me how you deal with it? \hl{I'm tired of having this every day} and feeling like I'll throw up.}}
    \caption{An example of stress being expressed in social media from our dataset, from a post in r/anxiety (reproduced exactly as found). Some possible expressions of stress are highlighted.}
    \label{fig:stress-example}
\end{figure}

\begin{table*}[ht!]
\centering
\begin{tabular}{|l|l|r|r|r|}
\hline
\multicolumn{1}{|c|}{\textbf{Domain}} & \multicolumn{1}{c|}{\textbf{Subreddit Name}} & \multicolumn{1}{c|}{\textbf{Total Posts}} & \multicolumn{1}{l|}{\textbf{Avg Tokens/Post}} & \multicolumn{1}{l|}{\textbf{Labeled Segments}} \\ \hline
\multirow{3}{*}{\textbf{abuse}}       & r/domesticviolence                      & 1,529                                     & 365                                           & 388                                            \\ \cline{2-5} 
                                      & r/survivorsofabuse                      & 1,372                                     & 444                                           & 315                                            \\ \cline{2-5} 
                                      & \textbf{Total}                          & \textbf{2,901}                                     & \textbf{402}                                           & \textbf{703}                                            \\ \hline
\multirow{3}{*}{\textbf{anxiety}}     & r/anxiety                               & 58,130                                    & 193                                           & 650                                            \\ \cline{2-5} 
                                      & r/stress                                & 1,078                                     & 107                                           & 78                                             \\ \cline{2-5} 
                                      & \textbf{Total}                          & \textbf{59,208}                                    & \textbf{191}                                           & \textbf{728}                                            \\ \hline
\multirow{5}{*}{\textbf{financial}}   & r/almosthomeless                        & 547                                       & 261                                           & 99                                             \\ \cline{2-5} 
                                      & r/assistance                            & 9,243                                     & 209                                           & 355                                            \\ \cline{2-5} 
                                      & r/food\_pantry                          & 343                                       & 187                                           & 43                                             \\ \cline{2-5} 
                                      & r/homeless                              & 2,384                                     & 143                                           & 220                                            \\ \cline{2-5} 
                                      & \textbf{Total}                          & \textbf{12,517}                                    & \textbf{198}                                           & \textbf{717}                                            \\ \hline
\textbf{PTSD}                         & r/ptsd                                  & 4,910                                     & 265                                           & 711                                            \\ \hline
\textbf{social}                       & r/relationships                         & 107,908                                   & 578                                           & 694                                            \\ \hline
\multicolumn{2}{|c|}{\textbf{All}}                                              & \textbf{187,444}                                   & \textbf{420}                                           & \textbf{3,553}                                          \\ \hline
\end{tabular}
\caption{\textbf{Data Statistics}. We include ten total subreddits from five domains in our dataset. Because some subreddits are more or less popular, the amount of data in each domain varies. We endeavor to label a comparable amount of data from each domain for training and testing.}
\label{tab:data-spread}
\end{table*}

Reddit is a social media website where users post in topic-specific communities called subreddits, and other users comment and vote on these posts. The lengthy nature of these posts makes Reddit an ideal source of data for studying the nuances of phenomena like stress. To collect expressions of stress, we select categories of subreddits where members are likely to discuss stressful topics:
\begin{itemize}
    \item \textbf{Interpersonal conflict}: abuse and social domains. Posters in the abuse subreddits are largely survivors of an abusive relationship or situation sharing stories and support, while posters in the social subreddit post about any difficulty in a relationship (often but not exclusively romantic) and seek advice for how to handle the situation.
    \item \textbf{Mental illness}: anxiety and Post-Traumatic Stress Disorder (PTSD) domains. Posters in these subreddits seek advice about coping with mental illness and its symptoms, share support and successes, seek diagnoses, and so on.
    \item \textbf{Financial need}: financial domain. Posters in the financial subreddits generally seek financial or material help from other posters.
\end{itemize}
We include ten subreddits in the five domains of abuse, social, anxiety, PTSD, and financial, as detailed in \autoref{tab:data-spread}, and our analysis focuses on the domain level. Using the {\tt PRAW} API,\footnote{\url{https://github.com/praw-dev/praw}} we scrape all available posts on these subreddits between January 1, 2017 and November 19, 2018; in total, 187,444 posts. As we will describe in \autoref{sec:annotation}, we assign binary stress labels to 3,553 segments of these posts to form a supervised and semi-supervised training set. An example segment is shown in \autoref{fig:stress-example}. Highlighted phrases are indicators that the writer is stressed: the writer mentions common physical symptoms (nausea), explicitly names fear and dread, and uses language indicating helplessness and help-seeking behavior.

The average length of a post in our dataset is 420 tokens, much longer than most microblog data (e.g., Twitter's character limit as of this writing is 280 characters). While we label segments that are about 100 tokens long, we still have much additional data from the author on which to draw. We feel this is important because, while our goal in this paper is to predict stress, having longer posts will ultimately allow more detailed study of the causes and effects of stress.

In \autoref{tab:data-examples}, we provide examples of labeled segments from the various domains in our dataset. The samples are fairly typical; the dataset contains mostly first-person narrative accounts of personal experiences and requests for assistance or advice. Our data displays a range of topics, language, and agreement levels among annotators, and we provide only a few examples. Lengthier examples are available in the appendix.

\begin{table*}[ht]
\centering
\begin{tabular}{|p{9.5cm}|c|c|c|}
\hline
\multicolumn{1}{|c|}{\textbf{Text}}                                                                                                                                                                                                                                                                                                                                                                                                                                                                                                                                                            & \multicolumn{1}{c|}{\textbf{Domain}} & \multicolumn{1}{c|}{\textbf{Label}} & \textbf{Ann. Agreed} \\ \hline
I only get it when I have a flashback or strong reaction to a trigger. I notice it sticks around even when I feel emotionally calm and can stick around for a long time after the trigger, like days or weeks. Its a new symptom I think. Also been having lots of nightmares again recently. Not sure what to do as I’m not currently in therapy, but I am waiting to be seen at a mental health clinic.                                                                                                                                                                                     & PTSD                                 & stress                              & 6/7 (86\%)                      \\ \hline
Regardless, that didn't last long, maybe half a year. I released that apartment, and most of my belongings (I kept a few boxes of my things from the military, personal effects, but little else). Looking back, there were some signs of emotional manipulation here, but it was subtle... and you know how it is, love is blind. We got engaged. It was quite the affair.                                                                                                                                                                                                                    & abuse                                & not stress                          & 5/5 (100\%)                     \\ \hline
Our dog Jett has been diagnosed with diabetes and is now in the hospital to stabilize his blood sugar. Luckily, he seems to be doing well and he will be home with us soon. Unfortunately, his bill is large enough that we just can't cover it on our own (especially with our poor financial situation). We're being evicted from our home soon and trying to find a place with this bill is just too much for us by ourselves. To help us pay the bill we've set up a GoFundMe. & financial                            & stress                              & 3/5 (60\%)                      \\ \hline
\end{tabular}
\caption{\textbf{Data Examples.} Examples from our dataset with their domains, assigned labels, and number of annotators who agreed on the majority label (reproduced exactly as found, except that a link to the GoFundMe has been removed in the last example). Annotators labeled these five-sentence segments of larger posts.}
\label{tab:data-examples}
\end{table*}

\subsection{Data Annotation} \label{sec:annotation}

We annotate a subset of the data using Amazon Mechanical Turk in order to begin exploring the characteristics of stress. We partition the posts into contiguous five-sentence chunks for labeling; we wish to annotate segments of the posts because we are ultimately interested in what parts of the post depict stress, but we find through manual inspection that some amount of context is important. Our posts, however, are quite long, and it would be difficult for annotators to read and annotate entire posts. This type of data will allow us in the future not only to \textit{classify} the presence of stress, but also to \textit{locate} its expressions in the text, even if they are diffused throughout the post.

We set up an annotation task in which English-speaking Mechanical Turk Workers are asked to label five randomly selected text segments (of five sentences each)  after taking a qualification test; Workers are allowed to select ``Stress'', ``Not Stress'', or ``Can't Tell'' for each segment. In our instructions, we define stress as follows:
``The Oxford English Dictionary defines stress as `a state of mental or emotional strain or tension resulting from adverse or demanding circumstances'. This means that stress results from someone being uncertain that they can handle some threatening situation. We are interested in cases where that someone also feels negatively about it (sometimes we can find an event stressful, but also find it exciting and positive, like a first date or an interview).''. We specifically ask Workers to decide whether the author is expressing both stress and a negative attitude about it, not whether the situation itself seems stressful. Our full instructions are available in the appendix.

We submit 4,000 segments, sampled equally from each domain and uniformly within domains, to Mechanical Turk to be annotated by at least five Workers each and include in each batch one of 50 ``check questions'' which have been previously verified by two in-house annotators. After removing annotations which failed the check questions, and data points for which at least half of the annotators selected ``Can't Tell'', we are left with 3,553 labeled data points from 2,929 different posts. We take the annotators' majority vote as the label for each segment and record the percentage of annotators who agreed. The resulting dataset is nearly balanced, with 52.3\% of the data (1,857 instances) labeled stressful.

Our agreement on all labeled data is $\kappa=0.47$, using Fleiss's Kappa \citep{fleiss}, considered ``moderate agreement'' by \citet{landis_koch_1977}. We observe that annotators achieved perfect agreement on 39\% of the data, and for another 32\% the majority was 3/5 or less.\footnote{It is possible for the majority to be less than 3/5 when more than 5 annotations were solicited.} 
This suggests that our data displays significant variation in how stress is expressed, which we explore in the next section.

\section{Data Analysis}

\begin{table*}[ht]
\centering
\begin{tabular}{|l|c|c|c|c|}
\hline
\multicolumn{1}{|c|}{\textbf{Domain}} & \textbf{``Negemo'' \%} & \multicolumn{1}{l|}{\textbf{``Negemo'' Coverage}} & \textbf{``Social'' \%} & \multicolumn{1}{l|}{\textbf{``Anxiety'' Coverage}} \\ \hline
\textbf{Abuse}                        & 2.96\%               & 39\%                                              & 12.03\%              & 58\%                                             \\ \hline
\textbf{Anxiety}                      & 3.42\%               & 37\%                                              & 6.76\%               & 62\%                                             \\ \hline
\textbf{Financial}                    & 1.54\%               & 31\%                                              & 8.06\%               & 42\%                                             \\ \hline
\textbf{PTSD}                         & 3.29\%               & 42\%                                              & 7.95\%               & 61\%                                             \\ \hline
\textbf{Social}                       & 2.36\%               & 38\%                                              & 13.21\%              & 59\%                                             \\ \hline
\textbf{All}                          & 2.71\%               & 62\%                                                & 9.62\%             & 81\%                                               \\ \hline
\end{tabular}
\caption{\textbf{LIWC Analysis by Domain.} Results from our analysis using LIWC word lists. Each term in quotations refers to a specific word list curated by LIWC; percentage refers to the percent of words in the domain that are included in that word list, and coverage refers to the percent of words in that word list which appear in the domain.}
\label{tab:domain-liwc}
\end{table*}

\begin{figure*}[ht!]
    \includegraphics[scale=0.4]{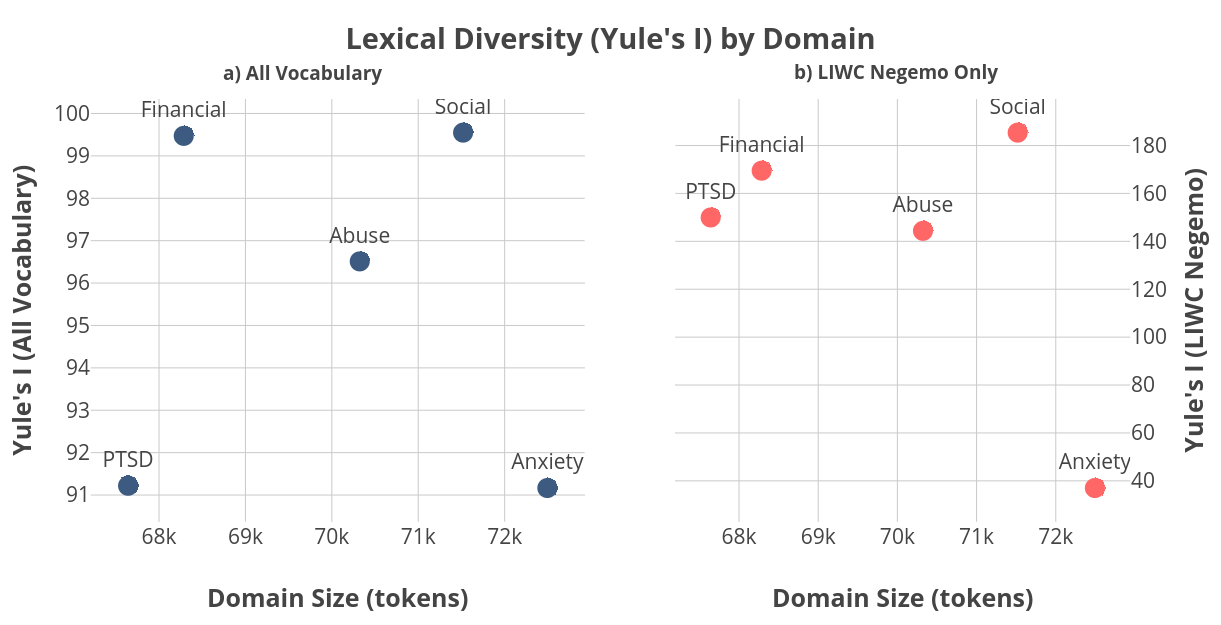}
    \centering
    \caption{\textbf{Lexical Diversity by Domain.} Yule's I measure (on the y-axes) is plotted against domain size (on the x-axes) and each domain is plotted as a point on two graphics. a) measures the lexical diversity of all words in the vocabulary, while b) deletes all words that were not included in LIWC's negative emotion word list.}
    \label{fig:domain-yule}
\end{figure*}

\begin{table*}[ht]
\centering
\begin{tabular}{|l|c|c|c|c|c|}
\hline
\multicolumn{1}{|c|}{\textbf{Label}} & \textbf{1st-Person \%} & \multicolumn{1}{l|}{\textbf{``Posemo'' \%}} & \textbf{``Negemo'' \%} & \multicolumn{1}{l|}{\textbf{``Anxiety'' Cover.}} & \multicolumn{1}{l|}{\textbf{``Social'' \%}} \\ \hline
\textbf{Stress}                      & 9.81\%                      & 1.77\%                                      & 3.54\%                 & 78\%                                               & 8.35\%                                      \\ \hline
\textbf{Non-Stress}                  & 6.53\%                       & 2.78\%                                      & 1.75\%                 & 67\%                                               & 11.15\%                                     \\ \hline
\end{tabular}
\caption{\textbf{LIWC Analysis by Label.} Results from our analysis using LIWC word lists, with the same definitions as in \autoref{tab:domain-liwc}. First-person pronouns (``1st-Person'') use the LIWC ``I'' word list.}
\label{tab:label-liwc}
\end{table*}

\begin{table}[ht]
\centering
\begin{tabular}{|c|r|r|}
\hline
\textbf{Measure} & \multicolumn{1}{c|}{\textbf{Stress}} & \multicolumn{1}{c|}{\textbf{Non-Stress}} \\ \hline
\% Conjunctions  & 0.88\%                               & 0.74\%                                   \\ \hline
Tokens/Segment   & 100.80                               & 93.39                                    \\ \hline
Clauses/Sentence & 4.86                                 & 4.33                                     \\ \hline
F-K Grade        & 5.31                                 & 5.60                                     \\ \hline
ARI              & 4.39                                 & 5.01                                     \\ \hline
\end{tabular}
\caption{\textbf{Complexity by Label.} Measures of syntactic complexity for stressful and non-stressful data.}
\label{tab:label-complexity}
\end{table}

\begin{figure}[ht]
    \includegraphics[scale=0.37]{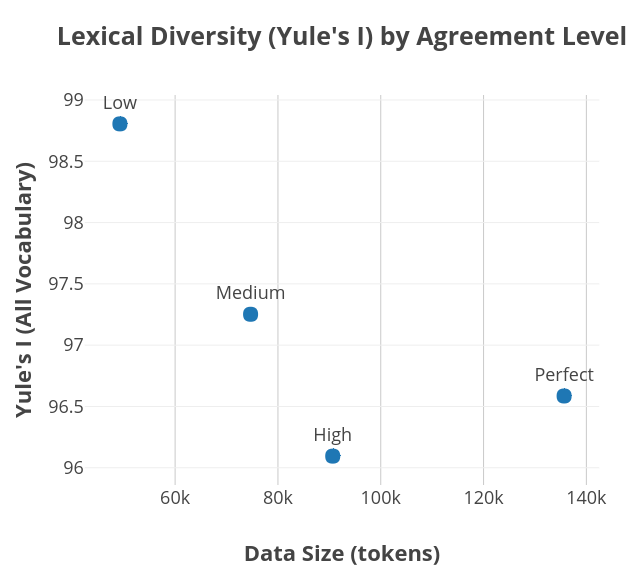}
    \centering
    \caption{\textbf{Lexical Diversity by Agreement.} Yule's I measure (on the y-axis) is plotted against domain size (on the x-axis) for each level of annotator agreement. Perfect means all annotators agreed; High, 4/5 or more; Medium, 3/5 or more; and Low, everything else.}
    \label{fig:agreement-diversity}
\end{figure}

While all our data has the same genre and personal narrative style, we find distinctions among domains with which classification systems must contend in order to perform well, and distinctions between stressful and non-stressful data which may be useful when developing such systems. Posters in each subreddit express stress, but we expect that their different functions and stressors lead to differences in how they do so in each subreddit, domain, and broad category.

\textbf{By domain.} We examine the vocabulary patterns of each domain on our training data only, not including unlabeled data so that we may extend our analysis to the label level. First, we use the word categories from the Linguistic Inquiry and Word Count (LIWC) \citep{liwc2015}, a lexicon-based tool that gives scores for psychologically relevant categories such as sadness or cognitive processes, as a proxy for topic prevalence and expression variety. We calculate both the percentage of tokens per domain which are included in a specific LIWC word list, and the percentage of words in a specific LIWC word list that appear in each domain (``coverage'' of the domain). 

Results of the analysis are highlighted in \autoref{tab:domain-liwc}. We first note that variety of expression
depends on domain and topic; for example, the variety in the expression of negative emotions is particularly low in the financial domain (with 1.54\% of words being negative emotion (``negemo'') words and only 31\% of ``negemo'' words used). We also see clear topic shifts among domains: the interpersonal domains contain roughly 1.5 times as many social words, proportionally, as the others; and domains are stratified by their coverage of the anxiety word list (with the most in the mental illness domains and the least in the financial domain).

We also examine the overall lexical diversity of each domain by calculating Yule's I measure \citep{yule_1944}. \autoref{fig:domain-yule} shows the lexical diversity of our data, both for all words in the vocabulary and for only words in LIWC's ``negemo'' word list. Yule's I measure reflects the repetitiveness of the data (as opposed to the broader coverage measured by our LIWC analysis). We notice exceptionally low lexical diversity for the mental illness domains, which we believe is due to the structured, clinical language surrounding mental illnesses. For example, posters in these domains discuss topics such as symptoms, medical care, and diagnoses (\autoref{fig:stress-example}, \autoref{tab:data-examples}). When we restrict our analysis to negative emotion words, this pattern persists only for anxiety; the PTSD domain has comparatively little lexical variety, but what it does have contributes to its variety of expression for negative emotions.

\textbf{By label.} We perform similar analyses on data labeled stressful or non-stressful by a majority of annotators. We confirm some common results in the mental health literature, including that stressful data uses more first-person pronouns (perhaps reflecting increased self-focus) and that non-stressful data uses more social words (perhaps reflecting a better social support network).

Additionally, we calculate measures of syntactic complexity, including the percentage of words that are conjunctions, average number of tokens per labeled segment, average number of clauses per sentence, Flesch-Kincaid Grade Level \citep{kincaid_fishburne_rogers_chissom_1975}, and Automated Readability Index \citep{ari}. These scores are comparable for all splits of our data; however, as shown in \autoref{tab:label-complexity}, we do see non-significant but persistent differences between stressful and non-stressful data, with stressful data being generally longer and more complex but also rated simpler by readability indices. These findings are intriguing and can be explored in future work.

\textbf{By agreement.} Finally, we examine 
the differences among annotator agreement levels. We find an inverse relationship between the lexical variety and the proportion of annotators who agree, as shown in \autoref{fig:agreement-diversity}. While the amount of data and lexical variety seem to be related, Yule's I measure controls for length, so we believe that this trend reflects a difference in the type of data that encourages high or low agreement.

\section{Methods}

In order to train supervised models, we group the labeled segments by post and randomly select 10\% of the posts ($\approx$ 10\% of the labeled segments) to form a test set. This ensures that while there is a reasonable distribution of labels and domains in the train and test set, the two do not explicitly share any of the same content. This results in a total of 2,838 train data points (51.6\% labeled stressful) and 715 test data points (52.4\% labeled stressful). Because our data is relatively small, we train our traditional supervised models with 10-fold cross-validation; for our neural models, we break off a further random 10\% of the training data for validation and average the predictions of 10 randomly-initialized trained models.

In addition to the words of the posts (both as bag-of-n-grams and distributed word embeddings), we include features in three categories:

\textbf{\textit{Lexical features.}} Average, maximum, and minimum scores for pleasantness, activation, and imagery from the Dictionary of Affect in Language (DAL) \citep{dal}; the full suite of 93 LIWC features; and sentiment calculated using the {\tt Pattern} sentiment library \citep{Smedt2012PatternFP}.

\textbf{\textit{Syntactic features.}} Part-of-speech unigrams and bigrams, the Flesch-Kincaid Grade Level, and the Automated Readability Index.

\textbf{\textit{Social media features.}} The UTC timestamp of the post; the ratio of upvotes to downvotes on the post, where an upvote roughly corresponds to a reaction of ``like'' and a downvote to ``dislike'' (\textbf{upvote ratio}); the net score of the post (\textbf{karma}) (calculated by Reddit, $n_\text{upvotes} - n_\text{downvotes}$)\footnote{\url{https://www.reddit.com/wiki/faq}}; and the total number of comments in the entire thread under the post.

\begin{table*}[ht!]
\centering
\begin{tabular}{|l|r|r|r|}
\hline
\multicolumn{1}{|c|}{\textbf{Model}} & \multicolumn{1}{c|}{\textbf{P}} & \multicolumn{1}{c|}{\textbf{R}} & \multicolumn{1}{c|}{\textbf{F}} \\ \hline
Majority baseline                    & 0.5161                          & 1.0000                          & 0.6808                          \\ \hline
CNN + features*                      & 0.6023                            & 0.8455                            & 0.7035                            \\ \hline
CNN*                                  & 0.5840                            & 0.9322                            & 0.7182                          \\ \hline
GRNN w/ attention + features*                & 0.6792                            & 0.7859                            & 0.7286                            \\ \hline
GRNN w/ attention*                           & 0.7020                            & 0.7724                            & 0.7355                          \\ \hline
n-gram baseline*                      & 0.7249                       & 0.7642                            & 0.7441                            \\ \hline
n-grams + features*                   & 0.7474                            & 0.7940                            & 0.7700                            \\ \hline
LogReg w/ pretrained Word2Vec + features  & 0.7346                            & 0.8103                            & 0.7706                            \\ \hline
LogReg w/ fine-tuned BERT LM + features*  & 0.7704                            & 0.8184                            & 0.7937                          \\ \hline
LogReg w/ domain Word2Vec + features*      & 0.7433                            & 0.8320                            & 0.7980                          \\ \hline
BERT-base*                           & 0.7518                             & 0.8699                             & 0.8065                          \\ \hline
\end{tabular}
\caption{\textbf{Supervised Results.} Precision (P), recall (R), and F1-score (F) for our supervised models. Our best model achieves 79.80 F1-score on our test set, comparable to the state-of-the-art pretrained BERT-base model. In this table, ``features'' always refers to our best-performing feature set ($\geq$ 0.4 absolute Pearson's $r$). Models marked with a * show a significant improvement over the majority baseline (approximate randomization test, $p < 0.01$).}
\label{tab:supervised-results}
\end{table*}

\subsection{Supervised Models}

We first experiment with a suite of non-neural models, including Support Vector Machines (SVMs), logistic regression, Na\"ive Bayes, Perceptron, and decision trees. We tune the parameters for these models using grid search and 10-fold cross-validation, and obtain results for different combinations of input and features. 

For input representation, we experiment with bag-of-n-grams (for $n \in \{1..3\}$), Google News pre-trained Word2Vec embeddings (300-dimensional) \citep{Mikolov:2013:DRW:2999792.2999959}, Word2Vec embeddings trained on our large unlabeled corpus (300-dimensional, to match), and BERT embeddings trained on our unlabeled corpus (768-dimensional, the top-level [CLS] embedding) \citep{devlin-etal-2019-bert}. We experiment with subsets of the above features, including separating the features by category (lexical, syntactic, social) and by magnitude of the Pearson correlation coefficient ($r$) with the training labels. Finally, we stratify the training data by annotator agreement, including separate experiments on only data for which all annotators agreed, data for which at least 4/5 annotators agreed, and so on.

We finally experiment with neural models, although our dataset is relatively small. We train both a two-layer bidirectional Gated Recurrent Neural Network (GRNN) \citep{DBLP:conf/emnlp/ChoMGBBSB14} and Convolutional Neural Network (CNN) (as designed in \citet{DBLP:journals/corr/Kim14f}) with parallel filters of size 2 and 3, as these have been shown to be effective in the literature on emotion detection in text (e.g., \citet{emo2vec,emonet}). Because neural models require large amounts of data, we do not cull the data by annotator agreement for these experiments and use all the labeled data we have. We experiment with training embeddings with random initialization as well as initializing with our domain-specific Word2Vec embeddings, and we also concatenate the best feature set from our non-neural experiments onto the representations after the recurrent and convolutional/pooling layers respectively.

Finally, we apply BERT directly to our task, fine-tuning the pretrained BERT-base\footnote{Using the implementation available at \url{https://github.com/huggingface/pytorch-transformers}} on our classification task for three epochs (as performed in \citet{devlin-etal-2019-bert} when applying BERT to any task). Our parameter settings for our various models are available in the appendix.

\section{Results and Discussion}\label{sec:results}

\begin{table*}[ht]
\centering
\begin{tabular}{ll|r|r|r|r|}
\cline{3-6}
\multicolumn{1}{c}{\textbf{}}                                & \multicolumn{1}{c|}{\textbf{}} & \multicolumn{4}{c|}{\textbf{Agreement Threshold for Data}}                          \\ \cline{3-6} 
\multicolumn{1}{c}{\textbf{}}                                & \multicolumn{1}{r|}{}          & Any Majority   & 60\% (3/5) & \multicolumn{1}{l|}{80\% (4/5)} & \multicolumn{1}{l|}{100\% (5/5)} \\ \hline
\multicolumn{1}{|l|}{\multirow{11}{*}{\textbf{Features}}} & None                    & 75.40 & 76.31     & 78.48                          & 77.69                          \\ \cline{2-6} 
\multicolumn{1}{|l|}{}                                       & All                   & 76.90 & 77.12     & 77.10                          & 78.28                          \\ \cline{2-6} 
\multicolumn{1}{|l|}{}                                       & LIWC                       & 77.91 & 78.91     & 78.16                          & 77.66                          \\ \cline{2-6} 
\multicolumn{1}{|l|}{}                                       & DAL                        & 75.58 & 77.06     & 78.05                          & 77.06                          \\ \cline{2-6} 
\multicolumn{1}{|l|}{}                                       & Lexical                    & 76.42 & 77.92     & 77.54                          & 77.88                          \\ \cline{2-6} 
\multicolumn{1}{|l|}{}                                       & Syntactic                  & 74.63 & 75.49     & 76.66                          & 76.19                          \\ \cline{2-6} 
\multicolumn{1}{|l|}{}                                       & Social                     & 76.67 & 76.45     & 78.38                          & 78.06                          \\ \cline{2-6} 
\multicolumn{1}{|l|}{}                                       & $|r| \geq$ 0.4        & 77.44 & 78.76     & \textbf{79.80}                 & 78.52                          \\ \cline{2-6} 
\multicolumn{1}{|l|}{}                                       & $|r| \geq$ 0.3        & 77.01 & 78.28     & 79.38                          & 78.31                          \\ \cline{2-6} 
\multicolumn{1}{|l|}{}                                       & $|r| \geq$ 0.2        & 77.53 & 78.61     & 79.02                          & 78.28                          \\ \cline{2-6} 
\multicolumn{1}{|l|}{}                                       & $|r| \geq$ 0.1        & 76.61 & 77.07     & 76.32                          & 77.48                          \\ \hline
\end{tabular}
\caption{\textbf{Feature Sets and Data Sets.} The results of our best classifier trained on different subsets of features and data. Features are grouped by type and by magnitude of their Pearson correlation with the train labels (no features had an absolute correlation greater than 0.5); data is separated by the proportion of annotators who agreed. Our best score (corresponding to our best non-neural model) is shown in bold.}
\label{tab:data-and-feat-comparison}
\end{table*}

We present our results in \autoref{tab:supervised-results}. Our best model is a logistic regression classifier with Word2Vec embeddings trained on our unlabeled corpus, high-correlation features ($\geq$ 0.4 absolute Pearson's $r$), and high-agreement data (at least 4/5 annotators agreed); this model achieves an F-score of 79.8 on our test set, a significant improvement over the majority baseline, the n-gram baseline, and the pre-trained embedding model, (all by the approximate randomization test, $p < 0.01$). The high-correlation features used by this model are LIWC's clout, tone, and ``I'' pronoun features, and we investigate the use of these features in the other model types. Particularly, we apply different architectures (GRNN and CNN) and different input representations (pretrained Word2Vec, domain-specific BERT). 

We find that our logistic regression classifier described above achieves comparable performance to BERT-base (approximate randomization test, $p > 0.5$) with the added benefits of increased interpretability and less intensive training. Additionally, domain-specific word embeddings trained on our unlabeled corpus  (Word2Vec, BERT) significantly outperform n-grams or pretrained embeddings, as expected, signaling the importance of domain knowledge in this problem.

We note that our basic deep learning models do not perform as well as our traditional supervised models or BERT, although they consistently, significantly outperform the majority baseline. We believe this is due to a serious lack of data; our labeled dataset is orders of magnitude smaller than neural models typically require to perform well. We expect that neural models can make good use of our large unlabeled dataset, which we plan to explore in future work. We believe that the superior performance of the pretrained BERT-base model (which uses no additional features) on our dataset supports this hypothesis as well.

In \autoref{tab:data-and-feat-comparison}, we examine the impact of different feature sets and levels of annotator agreement on our logistic regressor with domain-specific Word2Vec embeddings and find consistent patterns supporting this model. First, we see a tradeoff between data size and data quality, where lower-agreement data (which can be seen as lower-quality) results in worse performance, but the larger 80\% agreement data consistently outperforms the smaller perfect agreement data. Additionally, LIWC features consistently perform well while syntactic features consistently do not, and we see a trend towards the quality of features over their quantity; those with the highest Pearson correlation with the train set (which all happen to be LIWC features) outperform sets with lower correlations, which in turn outperform the set of all features. This suggests that stress detection is a highly lexical problem, and in particular, resources developed with psychological applications in mind, like LIWC, are very helpful.

\begin{table*}[]
\centering
\begin{minipage}{.3\linewidth}
    \centering
    \begin{tabular}{cc|c|c|}
    \cline{3-4}
                                                          &            & \multicolumn{2}{c|}{\textbf{Gold}} \\ \cline{3-4} 
                                                          & \textbf{}  & \textbf{0}       & \textbf{1}      \\ \hline
    \multicolumn{1}{|c|}{\multirow{2}{*}{\textbf{LogReg}}} & \textbf{0} & 241              & 105             \\ \cline{2-4} 
    \multicolumn{1}{|c|}{}                                 & \textbf{1} & 49               & 320             \\ \hline
    \end{tabular}
\end{minipage}
\begin{minipage}{.3\linewidth}
    \centering
    \begin{tabular}{cc|c|c|}
    \cline{3-4}
                                                          &            & \multicolumn{2}{c|}{\textbf{Gold}} \\ \cline{3-4} 
                                                          & \textbf{}  & \textbf{0}       & \textbf{1}      \\ \hline
    \multicolumn{1}{|c|}{\multirow{2}{*}{\textbf{BERT}}} & \textbf{0} & 240              & 106             \\ \cline{2-4} 
    \multicolumn{1}{|c|}{}                                 & \textbf{1} & 48               & 321             \\ \hline
    \end{tabular}
\end{minipage}
\begin{minipage}{.3\linewidth}
    \centering
    \begin{tabular}{cc|c|c|}
    \cline{3-4}
                                                          &            & \multicolumn{2}{c|}{\textbf{BERT}} \\ \cline{3-4} 
                                                          & \textbf{}  & \textbf{0}       & \textbf{1}      \\ \hline
    \multicolumn{1}{|c|}{\multirow{2}{*}{\textbf{LogReg}}} & \textbf{0} & 237              & 51             \\ \cline{2-4} 
    \multicolumn{1}{|c|}{}                                 & \textbf{1} & 53               & 374             \\ \hline
    \end{tabular}
\end{minipage}
\caption{\textbf{Confusion Matrices}. Confusion matrices of our best models and the gold labels. 0 represents data labeled not stressed while 1 represents data labeled stressed.}
\label{tab:confusion-matrices}
\end{table*}

\begin{table*}[ht]
\centering
\begin{tabular}{|>{\small}p{6.2cm}|c|c|c|c|}
\hline
\multicolumn{1}{|c|}{\thead{Text}} & \thead{Gold \\ Label} & \thead{Agreement} & \thead{Subreddit \\ Name} & \thead{Models \\ Failed} \\ \hline
Hello everyone, A very close friend of mine was in an accident a few years ago and deals with PTSD. He has horrific nightmares that wake him up and keep him in a state of fright. We live in separate provinces, so when he does have his dreams it is difficult to comfort him. Each time he calls, and I struggle with what to say on the phone. & Not Stress & 60\% & ptsd & Both \\ \hline
I asked the other day if they've set a date. He laughed in my face and said 'no' as if it were the most ridiculous thing he's ever heard. He comes home late, and showers immediately. Then, he showers every morning before he leaves. He doesn't talk to my mum and I, at all, and he's cagey and secretive about everything, to the point of hostility towards my sister. & Stress & 60\% & domesticviolence & BERT \\ \hline
If he's the textbook abuser, she is the textbook victim. She keeps giving him chances and accepting his apologies and living in this cycle of abuse. She thinks she's the one doing something wrong. I keep telling her that the only thing she is doing wrong is staying with this guy and thinking he will change. I tell her she does not deserve this treatment. & Not Stress & 100\% & domesticviolence & LogReg \\ \hline
\end{tabular}
\caption{\textbf{Error Analysis Examples}. Examples of test samples our models failed to classify correctly.``BERT'' refers to the state-of-the-art BERT-base model, while ``LogReg'' is our best logistic regressor described in \autoref{sec:results}.}
\label{tab:error-analysis-paper}
\end{table*}

Finally, we perform an error analysis of the two best-performing models. Although the dataset is nearly balanced, both BERT-base and our best logistic regression model greatly overclassify stress, as shown in \autoref{tab:confusion-matrices},
and they broadly overlap but do differ in their predictions (disagreeing with one another on approximately 100 instances).

We note that the examples misclassified by both models are often, though not always, ones with low annotator agreement (with the average percent agreement for misclassified examples being 0.55 for BERT and 0.61 for logistic regression). Both models seem to have trouble with less explicit expressions of stress, framing negative experiences in a positive or retrospective way, and stories where another person aside from the poster is the focus; these types of errors are difficult to capture with the features we used (primarily lexical), and further work should be aware of them. We include some examples of these errors in \autoref{tab:error-analysis-paper}, and further illustrative examples are available in the appendix.

\section{Conclusion and Future Work}

In this paper, we present a new dataset, Dreaddit, for stress classification in social media, and find the current baseline at 80\% F-score on the binary stress classification problem. We believe this dataset has the potential to spur development of sophisticated, interpretable models of psychological stress. 
Analysis of our data and our models shows that stress detection is a highly lexical problem benefitting from domain knowledge, but we note there is still room for improvement, especially in incorporating the framing and intentions of the writer.  We intend for our future work to use this dataset to contextualize stress and offer explanations using the content features of the text. Additional interesting problems applicable to this dataset include the development of effective distant labeling schemes, which is a significant first step to developing a quantitative model of stress. 

\section*{Acknowledgements}
We would like to thank Fei-Tzin Lee, Christopher Hidey, Diana Abagyan, and our anonymous reviewers for their insightful comments during the writing of this paper. This research was funded in part by a Presidential Fellowship from the Fu Foundation School of Engineering and Applied Science at Columbia University.

\bibliography{emnlp-ijcnlp-2019}

\begin{thebibliography}{27}
\expandafter\ifx\csname natexlab\endcsname\relax\def\natexlab#1{#1}\fi

\bibitem[{Abdul{-}Mageed and Ungar(2017)}]{emonet}
Muhammad Abdul{-}Mageed and Lyle~H. Ungar. 2017.
\newblock \href {https://doi.org/10.18653/v1/P17-1067} {Emonet: Fine-grained
  emotion detection with gated recurrent neural networks}.
\newblock In \emph{Proceedings of the 55th Annual Meeting of the Association
  for Computational Linguistics, {ACL} 2017, Vancouver, Canada, July 30 -
  August 4, Volume 1: Long Papers}, pages 718--728.

\bibitem[{Al-Shargie et~al.(2016)Al-Shargie, Kiguchi, Badruddin, Dass, and
  Hani}]{alshargie_kiguchi_badruddin_dass_hani_2016}
Fares Al-Shargie, Masashi Kiguchi, Nasreen Badruddin, Sarat~C. Dass, and Ahmad
  Fadzil~Mohammad Hani. 2016.
\newblock \href {https://doi.org/10.1364/BOE.7.003882} {Mental stress
  assessment using simultaneous measurement of eeg and fnirs}.
\newblock \emph{Biomedical Optics Express}, 7(10):3882–3898.

\bibitem[{Allen et~al.(2014)Allen, Kennedy, Cryan, Dinan, and
  Clarke}]{allen_kennedy_cryan_dinan_clarke_2014}
Andrew~P. Allen, Paul~J. Kennedy, John~F. Cryan, Timothy~G. Dinan, and Gerard
  Clarke. 2014.
\newblock \href {https://doi.org/10.1016/j.neubiorev.2013.11.005} {Biological
  and psychological markers of stress in humans: Focus on the trier social
  stress test}.
\newblock \emph{Neuroscience \& Biobehavioral Reviews}, 38:94–124.

\bibitem[{Calcia et~al.(2016)Calcia, Bonsall, Bloomfield, Selvaraj, Barichello,
  and Howes}]{calcia_bonsall_bloomfield_selvaraj_barichello_howes_2016}
Marilia~A. Calcia, David~R. Bonsall, Peter~S. Bloomfield, Sudhakar Selvaraj,
  Tatiana Barichello, and Oliver~D. Howes. 2016.
\newblock \href {https://doi.org/10.1007/s00213-016-4218-9} {Stress and
  neuroinflammation: a systematic review of the effects of stress on microglia
  and the implications for mental illness}.
\newblock \emph{Psychopharmacology}, 233(9):1637--1650.

\bibitem[{Cho et~al.(2014)Cho, van Merrienboer, G{\"{u}}l{\c{c}}ehre, Bahdanau,
  Bougares, Schwenk, and Bengio}]{DBLP:conf/emnlp/ChoMGBBSB14}
Kyunghyun Cho, Bart van Merrienboer, {\c{C}}aglar G{\"{u}}l{\c{c}}ehre, Dzmitry
  Bahdanau, Fethi Bougares, Holger Schwenk, and Yoshua Bengio. 2014.
\newblock \href {http://aclweb.org/anthology/D/D14/D14-1179.pdf} {Learning
  phrase representations using {RNN} encoder-decoder for statistical machine
  translation}.
\newblock In \emph{Proceedings of the 2014 Conference on Empirical Methods in
  Natural Language Processing, {EMNLP} 2014, October 25-29, 2014, Doha, Qatar,
  {A} meeting of SIGDAT, a Special Interest Group of the {ACL}}, pages
  1724--1734.

\bibitem[{Choudhury et~al.(2013)Choudhury, Gamon, Counts, and
  Horvitz}]{ICWSM136124}
Munmun~De Choudhury, Michael Gamon, Scott Counts, and Eric Horvitz. 2013.
\newblock \href
  {https://www.aaai.org/ocs/index.php/ICWSM/ICWSM13/paper/viewFile/6124/6351}
  {Predicting depression via social media}.
\newblock In \emph{Proceedings of the Seventh International AAAI Conference on
  Weblogs and Social Media}.

\bibitem[{Cohan et~al.(2018)Cohan, Desmet, Yates, Soldaini, MacAvaney, and
  Goharian}]{cohan-etal-2018-smhd}
Arman Cohan, Bart Desmet, Andrew Yates, Luca Soldaini, Sean MacAvaney, and
  Nazli Goharian. 2018.
\newblock \href {https://www.aclweb.org/anthology/C18-1126} {{SMHD}: a
  large-scale resource for exploring online language usage for multiple mental
  health conditions}.
\newblock In \emph{Proceedings of the 27th International Conference on
  Computational Linguistics}, pages 1485--1497, Santa Fe, New Mexico, USA.
  Association for Computational Linguistics.

\bibitem[{Devlin et~al.(2019)Devlin, Chang, Lee, and
  Toutanova}]{devlin-etal-2019-bert}
Jacob Devlin, Ming-Wei Chang, Kenton Lee, and Kristina Toutanova. 2019.
\newblock \href {https://www.aclweb.org/anthology/N19-1423} {{BERT}:
  Pre-training of deep bidirectional transformers for language understanding}.
\newblock In \emph{Proceedings of the 2019 Conference of the North {A}merican
  Chapter of the Association for Computational Linguistics: Human Language
  Technologies, Volume 1 (Long and Short Papers)}, pages 4171--4186,
  Minneapolis, Minnesota. Association for Computational Linguistics.

\bibitem[{Fleiss(1971)}]{fleiss}
Joseph~L Fleiss. 1971.
\newblock \href {https://doi.org/10.1037/h0031619} {Measuring nominal scale
  agreement among many raters}.
\newblock \emph{Psychological Bulletin}, 76(5):378--382.

\bibitem[{Guntuku et~al.(2018)Guntuku, Buffone, Jaidka, Eichstaedt, and
  Ungar}]{DBLP:journals/corr/abs-1811-07430}
Sharath~Chandra Guntuku, Anneke Buffone, Kokil Jaidka, Johannes~C. Eichstaedt,
  and Lyle~H. Ungar. 2018.
\newblock \href {http://arxiv.org/abs/1811.07430} {Understanding and measuring
  psychological stress using social media}.
\newblock \emph{CoRR}, abs/1811.07430.

\bibitem[{Kim(2014)}]{DBLP:journals/corr/Kim14f}
Yoon Kim. 2014.
\newblock \href {http://arxiv.org/abs/1408.5882} {Convolutional neural networks
  for sentence classification}.
\newblock \emph{CoRR}, abs/1408.5882.

\bibitem[{Kincaid et~al.(1975)Kincaid, Fishburne, Rogers, and
  Chissom}]{kincaid_fishburne_rogers_chissom_1975}
J.~Peter Kincaid, Robert~P. Fishburne, Richard~L. Rogers, and Brad~S. Chissom.
  1975.
\newblock \href {https://doi.org/10.21236/ada006655} {Derivation of new
  readability formulas (automated readability index, fog count and flesch
  reading ease formula) for navy enlisted personnel}.

\bibitem[{Kingma and Ba(2015)}]{DBLP:journals/corr/KingmaB14}
Diederik~P. Kingma and Jimmy Ba. 2015.
\newblock \href {http://arxiv.org/abs/1412.6980} {Adam: {A} method for
  stochastic optimization}.
\newblock In \emph{3rd International Conference on Learning Representations,
  {ICLR} 2015, San Diego, CA, USA, May 7-9, 2015, Conference Track
  Proceedings}.

\bibitem[{Landis and Koch(1977)}]{landis_koch_1977}
J.~Richard Landis and Gary~G. Koch. 1977.
\newblock \href {https://doi.org/10.2307/2529310} {The measurement of observer
  agreement for categorical data}.
\newblock \emph{Biometrics}, 33(1):159–174.

\bibitem[{Lin et~al.(2017)Lin, Jia, Qiu, Zhang, Shen, Xie, Tang, Feng, and
  Chua}]{7885098}
Huijie Lin, Jia Jia, Jiezhong Qiu, Yongfeng Zhang, Guangyao Shen, Lexing Xie,
  Jie Tang, Ling Feng, and Tat-Seng Chua. 2017.
\newblock \href {https://doi.org/10.1109/TKDE.2017.2686382} {Detecting stress
  based on social interactions in social networks}.
\newblock \emph{IEEE Transactions on Knowledge and Data Engineering},
  29(09):1820--1833.

\bibitem[{Lupien et~al.(2009)Lupien, McEwen, Gunnar, and
  Heim}]{lupien_mcewen_gunnar_heim_2009}
Sonia~J. Lupien, Bruce~S. McEwen, Megan~R. Gunnar, and Christine Heim. 2009.
\newblock \href {https://doi.org/10.1038/nrn2639} {Effects of stress throughout
  the lifespan on the brain, behaviour and cognition}.
\newblock \emph{Nature Reviews Neuroscience}, 10(6):434--445.

\bibitem[{Mikolov et~al.(2013)Mikolov, Sutskever, Chen, Corrado, and
  Dean}]{Mikolov:2013:DRW:2999792.2999959}
Tomas Mikolov, Ilya Sutskever, Kai Chen, Greg Corrado, and Jeffrey Dean. 2013.
\newblock \href {http://dl.acm.org/citation.cfm?id=2999792.2999959}
  {Distributed representations of words and phrases and their
  compositionality}.
\newblock In \emph{Proceedings of the 26th International Conference on Neural
  Information Processing Systems - Volume 2}, NIPS'13, pages 3111--3119, USA.
  Curran Associates Inc.

\bibitem[{Pedregosa et~al.(2011)Pedregosa, Varoquaux, Gramfort, Michel,
  Thirion, Grisel, Blondel, Prettenhofer, Weiss, Dubourg, Vanderplas, Passos,
  Cournapeau, Brucher, Perrot, and Duchesnay}]{scikit-learn}
F.~Pedregosa, G.~Varoquaux, A.~Gramfort, V.~Michel, B.~Thirion, O.~Grisel,
  M.~Blondel, P.~Prettenhofer, R.~Weiss, V.~Dubourg, J.~Vanderplas, A.~Passos,
  D.~Cournapeau, M.~Brucher, M.~Perrot, and E.~Duchesnay. 2011.
\newblock Scikit-learn: Machine learning in {P}ython.
\newblock \emph{Journal of Machine Learning Research}, 12:2825--2830.

\bibitem[{Pennebaker et~al.(2015)Pennebaker, Boyd, Jordan, and
  Blackburn}]{liwc2015}
James~W Pennebaker, Ryan~L Boyd, Kayla Jordan, and Kate Blackburn. 2015.
\newblock \href
  {https://repositories.lib.utexas.edu/bitstream/handle/2152/31333/LIWC2015_LanguageManual.pdf}
  {The development and psychometric properties of liwc2015}.

\bibitem[{Senter and Smith(1967)}]{ari}
R.J. Senter and E.A. Smith. 1967.
\newblock \href {https://apps.dtic.mil/dtic/tr/fulltext/u2/667273.pdf}
  {Automated readability index}.

\bibitem[{Smedt and Daelemans(2012)}]{Smedt2012PatternFP}
Tom~De Smedt and Walter Daelemans. 2012.
\newblock Pattern for python.
\newblock \emph{Journal of Machine Learning Research}, 13:2063--2067.

\bibitem[{Socher et~al.(2013)Socher, Perelygin, Wu, Chuang, Manning, Ng, and
  Potts}]{DBLP:conf/emnlp/SocherPWCMNP13}
Richard Socher, Alex Perelygin, Jean Wu, Jason Chuang, Christopher~D. Manning,
  Andrew~Y. Ng, and Christopher Potts. 2013.
\newblock \href {https://aclanthology.info/papers/D13-1170/d13-1170} {Recursive
  deep models for semantic compositionality over a sentiment treebank}.
\newblock In \emph{Proceedings of the 2013 Conference on Empirical Methods in
  Natural Language Processing, {EMNLP} 2013, 18-21 October 2013, Grand Hyatt
  Seattle, Seattle, Washington, USA, {A} meeting of SIGDAT, a Special Interest
  Group of the {ACL}}, pages 1631--1642.

\bibitem[{Whissel(2009)}]{dal}
Cynthia Whissel. 2009.
\newblock \href {https://doi.org/10.2466/PR0.105.2.509-521} {Using the revised
  dictionary of affect in language to quantify the emotional undertones of
  samples of natural language}.
\newblock \emph{Psychological Reports}, 105(2):509--521.

\bibitem[{Winata et~al.(2018)Winata, Kampman, and
  Fung}]{DBLP:journals/corr/abs-1805-12307}
Genta~Indra Winata, Onno~Pepijn Kampman, and Pascale Fung. 2018.
\newblock \href {http://arxiv.org/abs/1805.12307} {Attention-based {LSTM} for
  psychological stress detection from spoken language using distant
  supervision}.
\newblock \emph{CoRR}, abs/1805.12307.

\bibitem[{Xu et~al.(2018)Xu, Madotto, Wu, Park, and Fung}]{emo2vec}
Peng Xu, Andrea Madotto, Chien{-}Sheng Wu, Ji~Ho Park, and Pascale Fung. 2018.
\newblock \href {https://aclanthology.info/papers/W18-6243/w18-6243} {Emo2vec:
  Learning generalized emotion representation by multi-task training}.
\newblock In \emph{Proceedings of the 9th Workshop on Computational Approaches
  to Subjectivity, Sentiment and Social Media Analysis, WASSA@EMNLP 2018,
  Brussels, Belgium, October 31, 2018}, pages 292--298.

\bibitem[{Yule(1944)}]{yule_1944}
George~Udny Yule. 1944.
\newblock \emph{The statistical study of literary vocabulary}.
\newblock Cambridge Univ. Pr.

\bibitem[{Zuo et~al.(2012)Zuo, Li, and Fung}]{zuo-etal-2012-multilingual}
Xin Zuo, Tian Li, and Pascale Fung. 2012.
\newblock \href
  {http://www.lrec-conf.org/proceedings/lrec2012/pdf/594_Paper.pdf} {A
  multilingual natural stress emotion database}.
\newblock In \emph{Proceedings of the Eighth International Conference on
  Language Resources and Evaluation ({LREC}-2012)}, pages 1174--1178, Istanbul,
  Turkey. European Language Resources Association (ELRA).

\end{thebibliography}
\bibliographystyle{acl_natbib}

\clearpage
\appendix

\section{Data Samples}

\begin{figure*}
    \centering
    \fbox{\parbox{14cm}{Hey guys, I've been lurking around for a couple of weeks and I finally decided to post something.
    
    I'm not open to my parents about how I feel or what's going on with me because I fear that they won't understand or that I won't get my point across properly. The only person I can talk to is my SO.
    
    I'm a junior in high school, and I think I have anxiety. Although I'm not sure. I have this feeling of dread about school right before I go to bed and I wake up with an upset stomach which lasts all day and nakes me feel like I'll throw up. This causes me to lose appetite and not wanting to drink water for fear of throwing up. I'm not sure where else to go with this, but I need help.
    
    If any of you have this, can you tell me how you deal with it? I'm tired of having this every day and feeling like I'll throw up.
    
    If you guys need any follow up questions, feel free to ask and I'll do my best to answer.}}
    \caption{The full post for our example in \autoref{fig:stress-example}, posted in the subreddit r/anxiety.}
    \label{fig:data-appendix-1}
\end{figure*}

\begin{figure*}
    \centering
    \fbox{\parbox{14cm}{Does anybody else get this weird inner tremor when they’re really badly triggered? I can’t even see it on my hands but my body feels like it’s vibrating and if I put my teeth together they chatter slightly. 

I only get it when I have a flashback or strong reaction to a trigger. I notice it sticks around even when I feel emotionally calm and can stick around for a long time after the trigger, like days or weeks. 

It’s a new symptom I think. Also been having lots of nightmares again recently. Not sure what to do as I’m not currently in therapy, but I am waiting to be seen at a mental health clinic. }}
    \caption{The full post for one of our examples in \autoref{fig:stress-example}, posted in the subreddit r/ptsd.}
    \label{fig:data-appendix-2}
\end{figure*}

\begin{figure*}
    \centering
    \fbox{\parbox{14cm}{Why am I so alone? I have people all around me but knowing his presence is no longer around is terrifying me. 
    
Every other time we would fight or break up, he would stalk me. It use to frighten me he knew when I was on break if I spoke to someone or even what room I was at in my home. Now I find myself so lonely knowing that this is truly the end. I feel crazy for feeling this way but him and the kids were my LIFE for the last 2 years. 

I wasn't able to have social media and extremely restricted to whom I was allowed to talk to. Even those that were permitted were strongly monitored. 

Now not knowing the certainty of what the future holds is beyond terrifying. For the last 2 years my life was so controlled that I knew what I was to do minute by minute. I knew based on his moods how the day would present itself. If it was a "good day" I would be loved, cherished, showed with affection and promises. If it was a "bad day" I would be belittled, threatened, restricted from access to money and vehicle. On good days I would attempt to talk about the escalation and unacceptable behaviors displayed on bad days, however I could see the change in his eyes and snap everything would change. I was now the whore, the botch, the stupid incapable person he despises. Back to walking on eggshells. Trying to get that once perfect Cassanova back that once was there. But no even on good days the magic was gone, because it was now replaced with fear of the man I never knew before.
}}
    \caption{A full post from the subreddit r/domesticviolence.}
    \label{fig:data-appendix-3}
\end{figure*}

\begin{figure*}
    \centering
    \fbox{\parbox{14cm}{Our dog Jett has been diagnosed with diabetes and is now in the hospital to stabilize his blood sugar. Luckily, he seems to be doing well and he will be home with us soon. Unfortunately, his bill is large enough that we just can't cover it on our own (especially with our poor financial situation). 

We're being evicted from our home soon and trying to find a place with this bill is just too much for us by ourselves. [To help us with the bill, we set up a GoFundMe page] $<$URL$>$.

We need \$2,900 dollars to pay the bill in full, but any and all assistance is appreciated. Even just sharing would help us a lot. We've had Jett as part of our family for over 10 years and we want to make sure he gets better in a spot where we can to support him well. }}

    \caption{The full post for one of our examples in \autoref{tab:data-examples}, posted in the subreddit r/assistance. The brackets indicate a hyperlink.}
    \label{fig:data-appendix-4}
\end{figure*}

We include several full posts (with identifying information removed and whitespace collapsed) in \Cref{fig:data-appendix-1,fig:data-appendix-2,fig:data-appendix-3,fig:data-appendix-4}. Posts are otherwise reproduced exactly as obtained (with spelling errors, etc.). The selected examples are deliberately of a reasonable but fairly typical length for readability and space concerns; recall that our average post length is 420 tokens, longer for interpersonal subreddits and shorter for other subreddits. 

\section{Full Annotation Guidelines}

\begin{figure*}
    \includegraphics[scale=0.7]{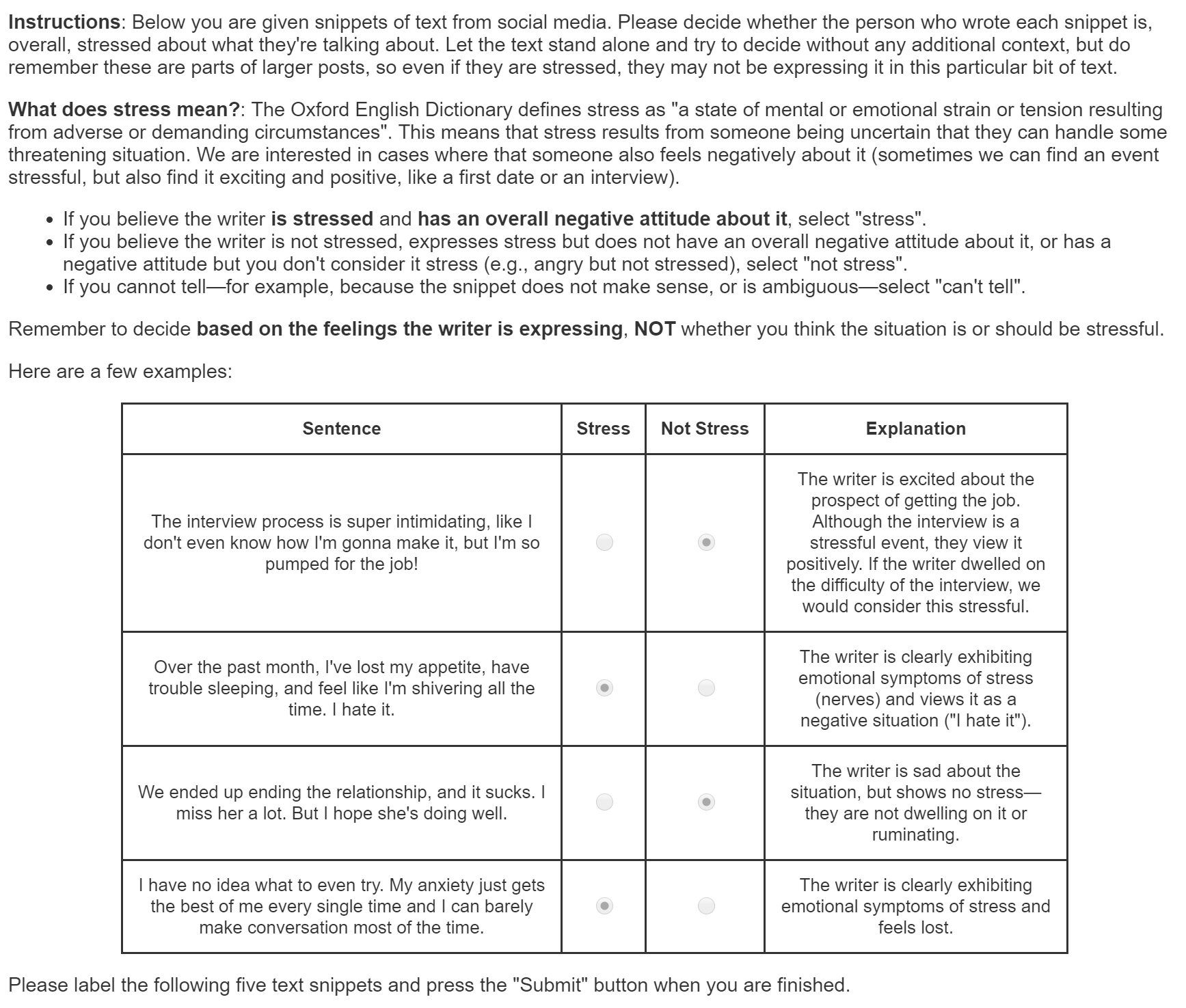}
    \caption{Our full annotation instructions.}
    \label{fig:annotation}
\end{figure*}

We provide our annotation instructions in full in \autoref{fig:annotation}. Mechanical Turk Workers were given these instructions and examples followed by five text segments (one of which was one of our 50 check questions) and allowed to select ``Stress'', ``Not Stress', or ``Can't Tell'' for each. Workers were given one hour to complete the HIT and paid \$0.12 for each HIT where they correctly answered the check question, with a limit of 30 total submissions per Worker.

\section{Parameter Settings}

We tune our traditional supervised models' parameters using grid search, all as implemented in Python's \texttt{scikit-learn} library \citep{scikit-learn}. Our best model uses unbalanced class weights, L2 penalty, and a constant term \texttt{C=10}, with other parameters at their default values. All cross-validation runs were initialized with the same random seed for comparability and reproducibility.

We train each of our neural models with the Adam optimizer \citep{DBLP:journals/corr/KingmaB14} for up to ten epochs with early stopping measured on the validation set. We apply a dropout rate of 0.5 during training in the recurrent layers and after the convolutional layers. We set our hidden sizes (i.e., the output of the recurrent and pooling layers) as well as our batch size to 128, and tune our learning rate to $5\cdot10^{-4}$; we set these parameters relatively small to try to work with our small data. We also experiment with scheduling the learning rate on plateau of the validation loss, and with pre-training the models on a much larger sentiment dataset, the Stanford Sentiment Treebank \citep{DBLP:conf/emnlp/SocherPWCMNP13}, to help combat the problem of small data, but this does not improve the performance of our neural networks.

\section{Error Analysis Examples}

\begin{table*}[ht]
\centering
\begin{tabular}{|>{\small}p{6.2cm}|c|c|c|c|}
\hline
\multicolumn{1}{|c|}{\thead{Text}} & \thead{Gold \\ Label} & \thead{Agreement} & \thead{Subreddit \\ Name} & \thead{Models \\ Failed} \\ \hline
We had 2 classes together, so we spent a few hours together most days working through problem sets. This next semester, I won't even have that. I'll probably be in more isolation this time around. Any tips are appreciated. Thanks! & Stress & 80\% & stress & Both \\ \hline
I developed and was diagnosed with PTSD 5 months later., I was having trouble sleeping (still kind of do), hypervigilant, moody and suicidal at times. I never thought I would make it through...but looking back,I used every single coping skill possible to survive, even if that meant calling crisis every day. I'm not perfect today but I really see the light at the end of the tunnel. I look forward to my future. & Not Stress & 57\% & ptsd & Both \\ \hline
She's the first person I've ever really opened up to. I haven't told her everything about whats happened, but she does know about my anxiety (which I get from my PTSD) and she reacts sportively to it. To some extent, I let me be "myself" around her, whatever I am. She's moving. She's moving to Maryland. & Stress & 80\% & ptsd & BERT \\ \hline
Sorry for the essay, poor grammar and punctuation. Thursday night. I asked a friend what they were up to tonight by text and instantly got a phone call after. Now usually I’d  ignore their called calls due to the worry of having an awkward phone conversation. This time I answer and we agree for them to come over. & Not Stress & 57\% & anxiety & BERT \\ \hline
We are in an uneasy peace right now, and i don't touch her although i am still the same caring guy but with precautions. What did i do to deserve this? and why do bad men get the good wife that sticks around and the good men always lose? I want to have my life back but i cannot see how? filling for a divorce will create a huge scandal since we are a minority as Christians and church laws prevail her, so i am also looking at a minimum of 6 years till any verdict is made. & Stress & 100\% & relationships & LogReg \\ \hline
\end{tabular}
\caption{\textbf{Additional Error Analysis Examples}. Supplementary examples for our error analysis.``BERT'' refers to the state-of-the-art BERT-base model, while ``LogReg'' is our best logistic regressor described in \autoref{sec:results}.}
\label{tab:error-analysis-appendix}
\end{table*}

As a supplement to our error analysis discussion in \autoref{sec:results}, we provide additional examples of test data points which one or both of our best models (BERT-base or our best logistic regressor with embeddings trained on our unlabeled corpus and high-correlation discrete features) failed to classify correctly in \autoref{tab:error-analysis-appendix}. 

\end{document}